\documentclass[runningheads]{llncs}
\usepackage{graphicx}
\usepackage{tikz}
\usepackage{comment}
\usepackage{amsmath,amssymb} % define this before the line numbering.
\usepackage{color}

\usepackage{graphicx}
\usepackage{amsmath}
\usepackage{amssymb}
\usepackage{booktabs}
\usepackage{multirow}
\newcommand{\cmark}{\ding{51}}%
\newcommand{\xmark}{\ding{55}}%
\usepackage{xcolor}
\newcommand{\fancomment}[1]{\textcolor[rgb]{1,0,0} {#1}}

\usepackage{comment}
\usepackage{amssymb}% http://ctan.org/pkg/amssymb
\usepackage{pifont}% http://ctan.org/pkg/pifont
\usepackage{bbm}

\newcommand{\ie}{\textit{i}.\textit{e}.}
\newcommand{\eg}{\textit{e}.\textit{g}.}

\newcommand\blfootnote[1]{%
  \begingroup
  \renewcommand\thefootnote{}\footnote{#1}%
  \addtocounter{footnote}{-1}%
  \endgroup
}

\usepackage{color,colortbl}
\definecolor{Gray1}{gray}{0.94}
\definecolor{Gray2}{gray}{0.99}

\usepackage[pagebackref,breaklinks,colorlinks]{hyperref}

\usepackage[capitalize]{cleveref}
\crefname{section}{Sec.}{Secs.}
\Crefname{section}{Section}{Sections}
\Crefname{table}{Table}{Tables}
\crefname{table}{Tab.}{Tabs.}

\usepackage[accsupp]{axessibility}  % Improves PDF readability for those with disabilities.

\begin{document}
% \linenumbers
\pagestyle{headings}
\mainmatter
\def\ECCVSubNumber{2666}  % Insert your submission number here

\title{Few-Shot Video Object Detection} % Replace with your title

\begin{comment}
\titlerunning{ECCV-22 submission ID \ECCVSubNumber} 
\authorrunning{ECCV-22 submission ID \ECCVSubNumber} 
\author{Anonymous ECCV submission}
\institute{Paper ID \ECCVSubNumber}
\end{comment}
%******************

%\begin{comment}
\titlerunning{Few-Shot Video Object Detection}
\author{Qi Fan\inst{1}\and %\thanks{This work was done when Qi was an intern at Kuaishou Technology.} \and %\orcidID{0000-1111-2222-3333} \and
Chi-Keung Tang\inst{1} \and %\orcidID{2222--3333-4444-5555}}
Yu-Wing Tai\inst{1,2}}
\authorrunning{Qi Fan et al.}
\institute{$^1$ The Hong Kong University of Science and Technology, 
%\email{fanqics@gmail.com, cktang@cs.ust.hk} \and
$^2$ Kuaishou Technology\\
\email{fanqics@gmail.com, cktang@cs.ust.hk, yuwing@gmail.com}}
%\end{comment}
%******************
\maketitle

%%%%%%%%% ABSTRACT
\begin{abstract}
We introduce Few-Shot Video Object Detection (FSVOD) with three contributions to real-world visual learning challenge in our highly diverse and dynamic world: 1) a large-scale video dataset FSVOD-500  comprising of 500 classes with class-balanced videos in each category for few-shot learning; 2) a novel Tube Proposal Network (TPN) to generate high-quality video tube proposals for aggregating feature representation for the target video object which can be highly dynamic; 3) a strategically improved Temporal Matching Network (TMN+) for matching representative query tube features with better discriminative ability thus achieving higher diversity.
Our TPN and TMN+ are jointly and end-to-end trained. Extensive experiments demonstrate that  our method produces significantly better detection results on two few-shot video object detection datasets compared to image-based methods and other naive video-based extensions. Codes and datasets are released at \url{https://github.com/fanq15/FewX}. %at \url{{https://github.com/fanq15/FewX}}.
\keywords{few-shot video object detection, %massive  video collections,
object indexing/retrieval,  tube proposal network, temporal matching network}\blfootnote{This research was supported in part by Kuaishou Technology, and the Research Grant Council of the Hong Kong SAR under grant No.~16201420.}
\end{abstract}

%%%%%%%%% BODY TEXT
\section{Introduction}

\begin{figure}[!t]
\centering
\includegraphics[width=1.0\linewidth]{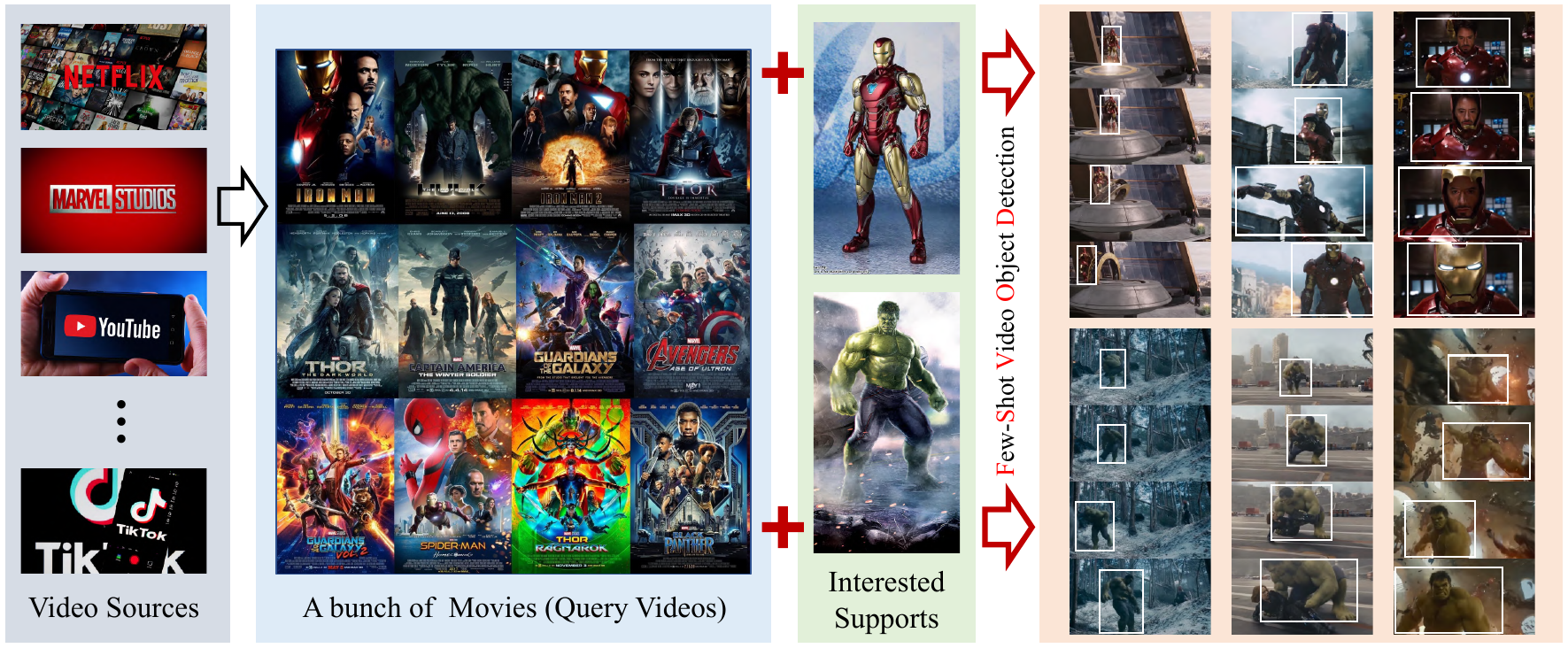}
%\fbox{\rule{0pt}{2.in} \rule{1.0\linewidth}{0pt}}
%\vspace{-0.1in}
\caption{Given only a few support objects of interest, our FSVOD detects all objects of the same category in query videos. Note that FSVOD enables object indexing/retrieval in a bunch of query videos to extract video clips containing the target objects.}
\label{fig:intro}
%\vspace{-0.2in}
\end{figure}

We ask the following question:
\textit{Given a bunch of videos, how can we index and localize all novel objects of interest as video clips?} See Figure~\ref{fig:intro}.

This problem is becoming increasingly essential 
with massive video collections in this media era: movies, YouTube videos, TikTok streaming videos, surveillance videos, just to name a few. 
The video objects of interests can be highly novel, often personalized, and thus are not covered by any existing datasets. Marvel fans may want to collect all Iron Man or Hulk clips from all Marvel movies, while warfare collectors want to create a TikTok  video consisting of tank clip collections from  war movies. We may not even know which 
videos contain the interested objects. % and we want to index these video clips for the target class.
%Although there are lots of such videos on Youtube or TikTok, keep in mind these videos are also created and uploaded by massive people.
%If we cannot suitably model this video processing problem, it will potentially takes up numerous time, labor and money.
%If we cannot suitably model this problem, it will wastes numerous time, labor and money for such video processing. 

%This problem is challenging.
%The video is arbitrary and massive. The interested object is arbitrary and massive. The video clips containing the target object is arbitrary and massive.
No existing tasks or solutions can solve this real-world challenge. Notably,
multiple object tracking~\cite{chu2019famnet,peng2020chained}, image/video object detection~\cite{girshick2015fast,ren2015faster,bertasius2018object,oh2019video} are all restricted in fixed and limited training classes.
Single object tracking~\cite{bertinetto2016fully,li2019siamrpn++} can track new classes, but it requires user-provided template for every video and can only track the target template object.
Few-shot learning seems a good candidate solution. But existing few-shot object detection~\cite{yang2020context,fan2020few} and few-shot classification~\cite{koch2015siamese,vinyals2016matching} are specifically designed for still images and they will produce numerous false positive results in videos. Few-shot video classification~\cite{cao2020few,zhu2018compound,khodadadeh2019unsupervised} does not target at instance recognition.

This real-world challenge motivates  few-shot video object detection (FSVOD): 
given only a few \emph{\textbf{support images}} %\emph{\textbf{support images}} 
of the target object in an unseen class, FSVOD detects all the objects belonging to the same class in a given \emph{\textbf{query video}}.
%Our FSVOD task suitably models this challenging problem.
The given support images can be arbitrary  objects of interest, and FSVOD works on arbitrary videos for indexing and localization.
%The rationale behind our successful problem modeling is as following:
The key to successful FSVOD is simultaneously modeling both high dynamics and high diversity  of our dynamic and diverse world, while other existing tasks can only contribute either high dynamic or high diversity, as summarized in  Table~\ref{table:task}, and thus falling short of the real-world challenge.

%Our contribution of FSVOD is built on previous significant tasks and greatly bridges the gap between researches and realistic problems for computer vision community.
The technical contributions of FSVOD, namely, Temporal Proposal Network (TPN) for high object dynamics and Temporal Matching Network (TMN+) for high object diversity, will be detailed. The core idea is to perform temporal matching between the tube-aggregated query features and supports, %Such design suitably fits our proposed task and problem. 
which enables high-quality detection based on the representative tube features and eliminates ghost objects (false positive predictions) which heavily suffers the few-shot image object detection methods.

The other contribution of this paper consists of a large-scale dataset that enables new research on few-shot video object detection. Our dataset contains 500 classes with a small and balanced number of high-quality videos in each class. 
%\fancomment{
The numerous classes with class-balanced videos enable the trained model to learn a general relation metric for novel classes. Note that this dataset contributes not only as the first benchmark for FSVOD, but also as a useful benchmark for other important vision tasks, such as multi-object tracking and video object detection which are still in lack of a well-constructed, class-balanced video benchmark on par in the number of classes as FSVOD-500.

\section{Related Work}

The FSVOD task is related to    few-shot learning, object detection and video understanding.  Table~\ref{table:task} summarizes its relationship with  closely related tasks. %computer vision tasks.

\begin{table}[!t]
\begin{center}
\tabcolsep=2pt
\caption{{\bf Comparing FSVOD and relevant computer vision tasks} in terms of {\em dyn}amic and {\em div}ersity capabilities: %modeling levels
%according to whether 
detecting {\em box} for {\em novel} object classes and/or {\em multiple} objects, and whether {\em temporal} information is considered. 
Number of `+' indicates how diversity each task can contribute. `S.A.' means scene adaptation.}
\label{table:task}
\vspace{-0.1in}
%{\footnotesize   %\normalsize
\begin{tabular}{cccccccc}
Task  & Dyn. & Div. & Box &  Nov. & Mul. & Temp. & S.A. \\
\hline

\hline
\cellcolor{Gray1} Image Object Detection (IOD)     & \cellcolor{Gray1} no & \cellcolor{Gray1} + & \cellcolor{Gray1} \cmark & \cellcolor{Gray1} \xmark & \cellcolor{Gray1} \cmark & \cellcolor{Gray1} \xmark & \cellcolor{Gray1} \cmark \\
\cellcolor{Gray2} Video Object Detection (VOD)       & \cellcolor{Gray2} yes & \cellcolor{Gray2} + & \cellcolor{Gray2} \cmark & \cellcolor{Gray2} \xmark & \cellcolor{Gray2} \cmark & \cellcolor{Gray2} \cmark & \cellcolor{Gray2} \cmark \\
%\hline
\cellcolor{Gray1} Multiple Object Tracking (MOT) 	           & \cellcolor{Gray1} yes & \cellcolor{Gray1} +  & \cellcolor{Gray1} \cmark & \cellcolor{Gray1} \xmark & \cellcolor{Gray1} \cmark & \cellcolor{Gray1} \cmark & \cellcolor{Gray1} \xmark \\
\cellcolor{Gray2} Single Object Tracking (SOT)              & \cellcolor{Gray2} yes & \cellcolor{Gray2} + & \cellcolor{Gray2} \cmark & \cellcolor{Gray2} \cmark & \cellcolor{Gray2} \xmark & \cellcolor{Gray2} \cmark & \cellcolor{Gray2} \xmark \\
\hline
\cellcolor{Gray1} Few-Shot Classification (FSC) 	        & \cellcolor{Gray1} no & \cellcolor{Gray1} ++ & \cellcolor{Gray1} \xmark & \cellcolor{Gray1} \cmark & \cellcolor{Gray1} \cmark & \cellcolor{Gray1} \xmark & \cellcolor{Gray1} \cmark \\
\cellcolor{Gray2} Few-Shot Object Detection (FSOD) 	        & \cellcolor{Gray2} no & \cellcolor{Gray2} +++ & \cellcolor{Gray2} \cmark & \cellcolor{Gray2} \cmark & \cellcolor{Gray2} \cmark & \cellcolor{Gray2} \xmark & \cellcolor{Gray2} -\cmark\\
\cellcolor{Gray1} {\bf Few-Shot Video Object Detection}   & \cellcolor{Gray1} yes & \cellcolor{Gray1} +++ & \cellcolor{Gray1} \cmark & \cellcolor{Gray1} \cmark & \cellcolor{Gray1} \cmark & \cellcolor{Gray1} \cmark & \cellcolor{Gray1} \cmark \\

\end{tabular}
%}
\end{center}
%\vspace{-0.05in}

%Number of `+'  indicates how much dynamics and diversity each task can contribute. 
%The '-' means negative and '+' means positive with different levels in terms of its number.
%\fancomment{ ``Novel'' means can detect novel classes, ``Multi.'' means can detect multiple objects, and ``Temp.'' means in the temporal domain.}
\vspace{-0.3in}
\end{table}

%\subsection{Few-Shot Learning}

\noindent{\bf Few-Shot Classification (FSC).~} 
%Few-shot learning has attracted many research studies. 
%Numerous works have contributed to few-shot classification. % with different methods. %
Optimization-based works learn task-agnostic knowledge on model parameters~\cite{Finn2017ModelAgnosticMF,bertinetto2018meta,lee2018gradient} %,gordon2018metalearning,lee2019meta,antoniou2018train,grant2018recasting,rusu2018meta} 
for fast adaptation to new tasks on limited training data, using only  a few gradient update steps. 
Some works~\cite{hariharan2017low,wang2018low} hallucinate new images for novel classes from limited labeled data. %with an image generator learned from base classes. %via data augmentation~\cite{}, synthesizing new data~\cite{} or using external data~\cite{}. 
Metric-based methods exploit a weight-shared network~\cite{koch2015siamese}
to extract features of the support and query images before feeding them to a transferable distance metric. %
Such matching strategy~\cite{vinyals2016matching,zhang2020deepemd,yang2018learning,snell2017prototypical} captures inherent variety between supports and queries  irrespective of classes and thus can be directly applied 
for classifying novel classes. %without any adaptation or fine-tuning. 
%Existing works have proposed enhanced  feature embedding with memory~\cite{santoro2016meta,cai2018memory}, local descriptors~\cite{li2019DN4,lifchitz2019dense}, class prototype representation~\cite{snell2017prototypical,oreshkin2018tadam}, or traversing  the entire support set~\cite{li2019finding}. Other works focus on learning an effective distance metric by exploiting parametric distance metric~\cite{yang2018learning}, Graph Neural Network (GNN)~\cite{kim2019edge,gidaris2019generating} or Earth Mover's Distance metric~\cite{zhang2020deepemd}. 

\noindent{\bf Few-Shot Object Detection (FSOD).~} With encouraging progress made in the few-shot classification, few-shot learning has continued to contribute to important computer vision tasks~\cite{dong2018few,Michaelis2018OneShotSI,Hu2019AttentionbasedMG,Gui2018FewShotHM,liu2020ppnet,li2020fss,fan2020cpmask} at a fast pace especially for object detection~\cite{yang2020context,perez2020incremental,wang2019meta}. 
In LSTD~\cite{Chen2018LSTDAL} %the authors propose to transfer knowledge from 
the gap between the source and target domain is minimized. 
%minimizing their gap of classification posterior probability.
RepMet~\cite{karlinsky2019repmet} learns the multi-modal distribution of the training classes in the embedding space. FR~\cite{kang2018few} exploits a meta feature learner %and a reweighting module 
to quickly adapt to novel classes. %Meta R-CNN~\cite{yan2019metarcnn} employs meta-learning over RoI features which turns the traditional object detection network into a meta-learner.   A two-stage fine-tuning approach is proposed in TFA~\cite{wang2020frustratingly} to detect novel classes. MPSR~\cite{wu2020multi} explores  multi-scale positive samples to refine  few-shot object detection prediction at various scales. FSDetView~\cite{xiao2020few} leverages  rich feature information from base classes to detect novel classes with a joint feature embedding module.
Some works exploit semantic relation reasoning~\cite{zhu2021semantic}, restore negative information~\cite{yang2020restoring}, feature hallucination~\cite{zhang2021hallucination} or other techniques~\cite{sun2021fsce,Hu_2021_CVPR,zhang2021accurate,Li_2021_CVPR,LiA_2021_CVPR,FanZB_2021_CVPR,LiBH_2021_CVPR,yan2019metarcnn,wang2020frustratingly,wu2020multi,xiao2020few} to facilitate few-shot object detection.
All of the above methods however require fine-tuning on novel classes. 
In FSOD~\cite{fan2020few} the authors proposed to learn a matching metric with attention RPN and multi-relation detector to detect novel classes. %without fine-tuning. 

%\fancomment{
Our FSVOD extends FSOD task to the temporal domain, with the technical approach motivated by the matching network~\cite{vinyals2016matching} and FSOD network~\cite{fan2020few} to detect novel classes without fine-tuning.
%}

%\subsection{Object Detection}

\noindent{\bf Image Object Detection (IOD).~}
Existing object detection methods can be mainly categorized to the two-stage approach~\cite{girshick2015fast,ren2015faster,lin2017feature} and one-stage approach~\cite{lin2017focal,redmon2016you,redmon2017yolo9000,liu2016ssd,liu2018receptive,zhang2020bridging}, based on whether  a region-of-interest proposal step is used. The two-stage approach was pioneered by R-CNN~\cite{girshick2014rich}.
%which extracts region features from vast number of proposals before feeding them into a second stage to classify the classes and refine the locations. 
In recent years, this approach has been improved by various excellent works
% shrivastava2016contextual,qin2019thundernet,najibi2016g,shrivastava2016training,he2019bounding
and achieved remarkable performance~\cite{he2017mask,singh2018sniper,cai2016unified,cai2018cascade,dai2016r,li2019scale,bell2016inside}. The one-stage approach on the other hand discards the proposal generation procedure in lieu of higher computational efficiency and faster inference speed with anchor-based~\cite{kong2017ron,shen2017dsod,zhu2019scratchdet,zhang2018single} %,zhangsf2018single}%,wang2019learning,nie2019enriched,tan2020efficientdet} 
or anchor-free detectors~\cite{law2018cornernet,lu2019grid,zhou2019bottom,duan2019centernet,yang2019reppoints,tian2019fcos,liu2019high,kong2020foveabox}. %This approach can be further categorized to anchor-based and anchor-free detectors, based on whether the pertinent method tiles a large number of preset anchors~\cite{kong2017ron,shen2017dsod,zhu2019scratchdet,zhang2018single,zhangsf2018single,wang2019learning,nie2019enriched,tan2020efficientdet} or directly finds objects from keypoints~\cite{law2018cornernet,lu2019grid,zhou2019bottom,duan2019centernet,yang2019reppoints} or center points~\cite{tian2019fcos,liu2019high,kong2020foveabox}.

\noindent{\bf Video Object Detection (VOD).}
Video object detection aims at detecting objects of pre-defined classes in a given video. 
%Although object detection methods have achieved significant success in still images, it is not trivial to generalize them to the video domain due to the challenges arisen from scene dynamics.%, e.g., motion blur, video defocus, and illumination differences. 
%Numerous algorithms focus on improving per-frame object results by exploiting temporal information.  
Some  enhance the quality of per-frame features by integrating temporal information locally~\cite{bertasius2018object,deng2019relation,xiao2018video,wang2018fully},  globally~\cite{deng2019object,shvets2019leveraging,wu2019sequence} or both~\cite{oh2019video,woo2018linknet,wu2019long,xu2019spatial,chen2020memory}, while others follow the ``sequential detection tracking" paradigm~\cite{feichtenhofer2017detect,zhu2018towards,zhu2017flow,zhu2017deep,kang2017object,tang2019object} to associate and rescore  detected boxes on individual frames. The above work in intensive supervision and cannot be applied readily to detect novel classes.
VOD variants include
%There are also many subtasks by adding different other requirements to MOT, 
e.g., video object segmentation (VOS)~\cite{Perazzi2016,xu2018youtube}, video instance segmentation (VIS)~\cite{yang2019video} and video panoptic segmentation (VPS)~\cite{kim2020video}.
%%\vspace{0.05in}
%\fancomment{

Both IOD and VOD are restricted to  pre-defined classes making it hard for them to detect novel classes. FSVOD eliminates this restriction with its detection generality on novel classes in videos. 
%However, given their close relationship, we still adopt the same evaluation metric to evaluate the detection performance on individual frames.
%}

%\subsection{Object Tracking}

\noindent{\bf Single Object Tracking (SOT).}
Given an arbitrary target with its location in the first frame, single object tracking aims to infer its location in  subsequent frames of the given video. Thanks to the construction of new benchmark datasets~\cite{fan2019lasot,wu2013online} and annually held tracking challenges~\cite{kristan2015visual,kristan2017visual,kristan2018sixth}, we have witnessed rapid performance boost in the last decade. The correlation filter based trackers~\cite{danelljan2017eco,danelljan2015learning,danelljan2014adaptive,henriques2014high} achieve superb performance with efficient inference speed. 
%by transforming the object template matching problem into a correlation operation in the frequency domain. 
The recent emerging siamese network based trackers~\cite{bertinetto2016fully,guo2017learning,li2018high,li2019siamrpn++,held2016learning,tao2016siamese,valmadre2017end} %,wang2018learning}
%%,zhu2018distractor} 
have drawn much attention due to their well-balanced performance and efficiency. 
%, which perform tracking by learning a general similarity map between the feature representations of the target template and the search region.

Although SOT models can track unseen objects, they heavily rely on the provided template and can only track one target object. 
The online tracking trackers~\cite{bhat2019learning,goutam2020learning,danelljan2020probabilistic,bhat2020know,danelljan2014adaptive,danelljan2017eco,danelljan2015learning,atom2019,keeptrack} can be finetuned/updated on the first frame, but they focus on tracking single object
with the video-specific annotated first frame. On other hand,
our FSVOD focuses on detecting arbitrary novel objects in
videos based on given video-agnostic support images even
from other images/videos and can be reused for all input
videos.

\noindent{\bf Multiple Object Tracking (MOT).~}
%Multiple video object tracking is a fundamentally challenging problem:
This task~\cite{valmadre2021local,luiten2021hota} requires simultaneous prediction of spatio-temporal location and classification of video objects into pre-defined classes. Current mainstream trackers~\cite{bergmann2019tracking,lu2020retinatrack,chu2019famnet,zhan2020simple,zhou2020tracking,kim2015multiple,sadeghian2017tracking,tang2017multiple,fang2018recurrent,yu2016poi} 
adopt  tracking-by-detection (TBD) by first performing per-frame detection and then associating the detected boxes in the temporal dimension. Some works  leverage trajectories or tubes to capture motion trails of targets~\cite{peng2020chained,kang2016object,shao2018find,zhu2018online,pang2020tubetk}. 
%multi-object tracking and segmentation (MOTS)~\cite{voigtlaender2019mots}
%to facilitate the tracking task. 

%\fancomment{
While MOT models can simultaneously track multiple objects, they cannot generalize
to novel classes. FSVOD can detect novel classes in videos. Our technical approach is inspired by these previous methods, especially tube-based MOT and VOD methods, \eg, CPN~\cite{tang2019object} and CTracker~\cite{peng2020chained}, which are restricted in limited training classes.
%}

\section{Proposed Method}
%\subsection{Problem Definition}

%Different from traditional video object detection which requires the model to detect predefined seen classes, 
Few-shot video object detection aims at  detecting \emph{\textbf{novel}} classes unseen in the training set. Given a \emph{\textbf{support image}} containing one object of the support class $c$ and a \emph{\textbf{query video}} sequence with $T$ frames, the task is to detect all the objects belonging to the support class $c$ in every frame. Suppose the support set contains $N$ classes with $K$ samples for each class, the problem is defined as $N$-way $K$-shot detection. 
Specifically, during inference, if all the support classes are exploited for detection, it is dubbed full-way evaluation.

%{\bf CK: some of the above symbols such as $s_c$ and $v$ are never used; remove all symbols that are never used. }
% Removed.

\subsection{Overview}
Technically, it is non-trivial to transfer few-shot learning~\cite{koch2015siamese,santoro2016meta,Finn2017ModelAgnosticMF} %,lifchitz2019dense,snell2017prototypical,oreshkin2018tadam,yang2018learning}
to the video object detection domain for simultaneously modeling the dynamic and diverse world. Few-shot learning requires a large-scale, class-balanced dataset with numerous base classes to train a class-agnostic metric capable of generalizing to novel classes~\cite{sbai2020impact,fan2020few,li2020fss}. Besides, videos present additional data challenges  caused by \eg, motion blur, occlusion and deformation of objects, making  infeasible straightforward extension of few-shot image  to few-shot video object detection without adequate temporal consideration.  

This paper extends 
%The  first contribution consists of the novel extension of 
the traditional video object detection to detect novel classes in a few-shot learning setting which is not a straightforward problem. We propose a novel tube-based few-shot video object detection model for detecting novel classes in a given video, without any fine-tuning or retraining.
%This opens up possibilities and applications that require detection of novel classes in  videos.
We make the following contributions:
%To address the above issues, this paper makes the following three contributions:
%we construct a large-scale dataset with vast classes (500 classes), and propose a tube-based few-shot learning model in the video domain.

%The key point of few-shot video object detection is to use only a few given object images to detect the objects with the same class in the video sequence. Although few-shot object detection methods provide good solutions in the image field, it is hard for these image-based methods to solve the video detection problem without involving the essential temporal information. Objects in the adjacent frames share similar spatial location and appearance with small deformation. And we believe that the temporal similarity can be well leveraged to detect novel classes in the video. This paper has three contributions: 

%We present detailed  investigation and comparison  in the paper.
%the novel task and compare to other tasks for better understanding of this task.
%\textcolor{red}{The technical contribution is too long?}
%Our first contribution is a novel tube-based few-shot video object detection model for detecting novel classes in a given video, without any fine-tuning or retraining. 
We first model {\it dynamic objects} by generating temporal tubes using our novel Tube Proposal Network (TPN) exploiting spatial adjacency and appearance similarity in the neighboring frames. Specifically, by introducing  novel inter-frame proposals to detect objects in consecutive frames, %temporal continuity of objects across frames, 
TPN can capture  potential objects in the query video while filtering out background and ghost objects (the false positive objects detected in isolated frames). We argue that the aggregated features
across frames can better represent the target objects which leads to significant improvement on the detection performance. %in videos.

%The second contribution consists of the 
Then we model {\it diversity of objects} using subsequent Temporal Matching Network (TMN+), which is specially designed and strategically improved to match support features and the aggregated query features from temporal tube proposals generated by TPN. Our proposed TMN+ effectively leverages %takes full advantage of 
the representative tube features by bridging the gap between training and inference via our novel temporal alignment branch. Furthermore, a new support classification loss is used to learn a highly discriminative feature, and a label-smoothing regularization is used for better generalization on novel unseen classes. Consequently, our TMN+ boosts matching performance on novel classes without extra computation overhead at inference. 

\begin{figure}[!t]
\centering
    \includegraphics[width=0.9\textwidth]{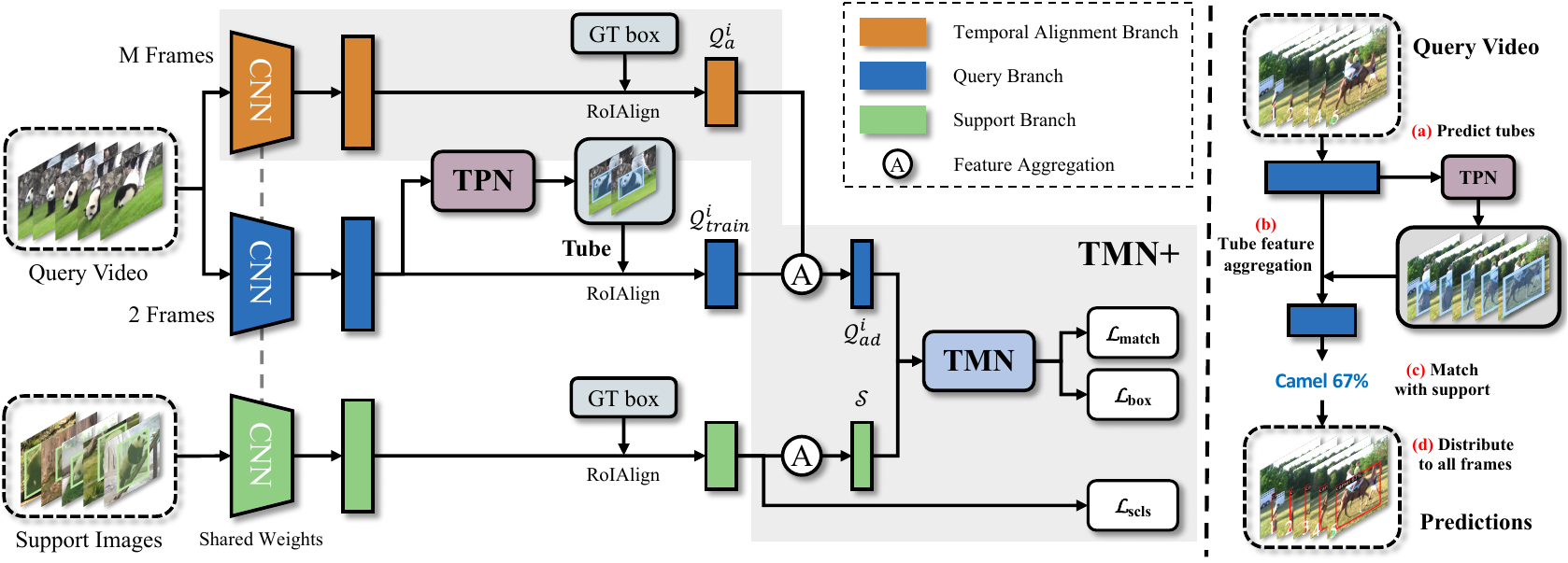}
%\fbox{\rule{0pt}{2.in} \rule{1.0\linewidth}{0pt}}
\vspace{-0.1in}
    \caption{
    {\bf Network architecture} at training (left) and testing (right) stages. The query video and support images are processed by the weight-shared backbone. The query branch only processes two query images. The temporal alignment branch (TAB) is used for query feature alignment, and a classification module is introduced to produce representative support features. For clarity we show the detection on a single object, while our  model can perform multi-object detection with corresponding tubes.} % In the inference stage, the TPN generates tubes for each potential object in the query video. The aggregated features are fed to TMN to match with support features. In the training stage, 
\label{fig:network}
\vspace{-0.2in}
\end{figure}

The TPN and TMN+ are integrated into one unified network and jointly optimized in an end-to-end manner to simultaneously handle high dynamics and diversity in visual object detection.

%Furthermore, we introduce a support classification loss to learn a highly discriminative feature, which can be generalized from the trained seen classes to  novel unseen classes. Finally, we enable the multi-relation head to analyze temporal relationship and cooperate with the temporal tubes generated by the TPN.

%500 classes with only a small number of videos in each class thus making it easily extensible. 

%\fancomment{And we hope the new FSVOD task, dataset and algorithm can help us to step forward towards video understanding in the real-world.}

%\textcolor{red}{Naive Solutions (Here or introduction?)}

%State the disadvantages of other naive solutions: FSOD-based methods, IOD-based methods, VOD-based methods and MOT-based methods. Then introduce our method.

\subsection{Few-Shot Video Object Detection Network}

Figure~\ref{fig:network} shows the network architecture. We propose a novel temporal detection network that %to perform few-shot object detection in videos. Our key idea is to 
exploits  tubes to locate and represent objects in the temporal domain, which are then matched with support features.   %(Maybe we need a more interesting version.) 

\vspace{-0.2in}

\subsubsection{Tube Proposal Network}

In image object detection, region proposal network RPN~\cite{ren2015faster} has become a classical module to produce proposals for potential objects while filtering out the background. These proposals are fed to the R-CNN head for finer classification and localization. %RPN  produces proposals in the spatial domain and thus unsuitable in  spatio-temporal processing. Thus, 

\begin{figure}[!t]
\centering
%\fbox{\rule{0pt}{2.5in} \rule{1.0\linewidth}{0pt}}
%0.845
\includegraphics[width=0.9\linewidth]{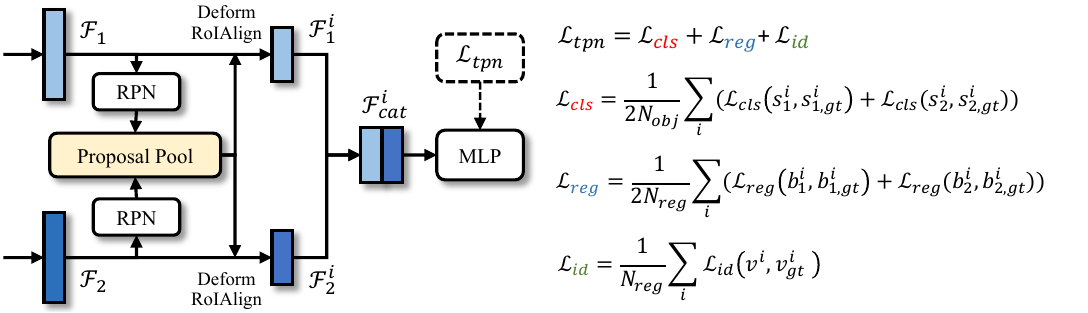}
\vspace{-0.1in}
\caption{Tube Proposal Network (TPN) and the loss function. The $*_{gt}$ is the ground-truth label for the corresponding prediction, $\mathcal{L}_{\text{cls}}$ and $\mathcal{L}_{\text{id}}$ are both cross-entropy loss and $\mathcal{L}_{\text{reg}}$ is the smooth $\mathcal{L}_1$ loss. 
$N_{obj}$ and $N_{reg}$ are respectively the number of proposals and foreground proposals.}
\label{fig:tpn}
\vspace{-0.2in}
\end{figure}

We extend RPN to the temporal domain to generate \emph{\textbf{tube proposals}} to locate and represent objects across frames. %In this way, the model can exploit not only the image appearance in the current frame but also temporal information from adjacent frames.
The resulting network is our novel tube proposal network (Figure~\ref{fig:tpn}) which exploits the high likelihood that %general both the location and appearance clues in videos where 
the same object in neighboring frames tend to have \emph{\textbf{similar location and  appearance}}.

%\noindent{\bf Location.} 
To utilize the location cue in adjacent frames, we propose the novel \emph{\textbf{ inter-frame proposals}} by feeding the same proposals to two adjacent frames. Note that proposals usually serve as a coarse prediction prior for later finer regression. The predicted boxes regressed from the same proposals indicate the same objects and therefore  inter-frame proposals can associate objects across frames.
However, it is also possible that objects with large motion may locate far away in adjacent frames, or the locations are occupied by other objects in the next frame. 
To address this problem, we adopt the deformable RoIAlign~\cite{dai2017deformable} operator to enlarge the search region for the target objects by adapting the sampling bins conditioned on the input feature.
%\vspace{5pt}
%\vspace{-0.1in}
%\noindent {\bf Appearance. } 
To exploit the appearance cue in neighboring frames to address the second problem, we verify the same object by predicting the identification score of the predicted boxes regressed from the same proposal.

%\vspace{0.05in}

%\vspace{-0.1in}

Specifically, %(see Figure~\ref{fig:tpn}), 
%for the potential instance $i$,
given two adjacent frames $\{I_1, I_2\}$, 
we first use RPN to generate proposals for each frame and collect both frame proposals to construct the proposal pool.
%we first extract features $\{\mathcal{F}_t, \mathcal{F}_{t+1}\}$, and use RPN to generate proposals for each frame. %to construct the proposal pool $P$.
Each proposal $p_i$ in the proposal pool is simultaneously fed to the two frames to extract proposal features $\{\mathcal{F}_1^i, \mathcal{F}_2^i\}$ with the deformable RoIAlign operator. These proposal features from individual frames are concatenated as $\mathcal{F}_{cat}^i=\mbox{concat}(\mathcal{F}_1^i, \mathcal{F}_2^i)$, which is then fed to the following multilayer perceptron (MLP) layer to perform objectness classification $\{s_1^i, s^i_2\}$, box regression $\{b_1^i, b^i_2\}$ for each frame, and identify verification $v^i$.
The 2-frame tube prediction is trained with the TPN loss $\mathcal{L}_{tpn}$, as shown in Figure~\ref{fig:tpn}.

\begin{comment}

%\vspace{-0.3in}
\begin{equation}
\begin{aligned}
    &\mathcal{L}_{\text{tpn}} = \frac{1}{2N_{obj}} \sum_i (\mathcal{L}_{\text{cls}}(s_t^i, s_{t,gt}^i) + \mathcal{L}_{\text{cls}}(s_{t+1}^i, s_{t+1, gt}^i)) \\ 
     & +\frac{1}{2N_{reg}} \sum_i (\mathcal{L}_{\text{reg}}(b_t^i, b_{t,gt}^i)  + \mathcal{L}_{\text{reg}}(b_{t+1}^i, b_{t+1,gt}^i)) \\
     & + \frac{1}{N_{reg}} \sum_i \mathcal{L}_{\text{id}}(v^i, v_{gt}^i), 
\end{aligned}
\end{equation}

\end{comment}

%\noindent{\bf Inference.}
During inference, the TPN needs to connect all frames in the given video by repeating the 2-frame tube prediction. Consider the 3-frame case where the $T$-frame ($T>3$) can be generalized~\footnote{The operations are parallel conducted for each instance/track $i$ and we omit the instance notion for simplicity.}: given $\{I_1, I_2, I_3\}$, we first send {\color{blue}$\{I_1, I_2\}$} to the model to generate a 2-frame tube {\color{blue}$\{b_1, b_2\}$}. Then we feed the pre-computed tube box {\color{blue}$b_2$} %in the last frame $I_t$ 
to $\{{\color{blue}I_2}, {\color{red}I_3}\}$ as the inter-frame proposal to generate the tube box {\color{red}$b_3$} for frame {\color{red}$I_3$} to construct another 2-frame tube $\{{\color{blue}b_2}, {\color{red}b_3}\}$.
%Note that the inter-frame proposal {\color{blue}$b_2$} is pre-computed in the previous tube.
We can construct a 3-frame tube $\{{\color{blue}b_1, b_2}, {\color{red}b_3}\}$ by linking {\color{blue}$\{b_1, b_2\}$} and $\{{\color{blue}b_2}, {\color{red}b_3}\}$ through the inter-frame proposal {\color{blue}$b_2$}. The overlapped frame $I_2$ is used to verify the same objects between two frame pairs and its feature are reused in the process to avoid  repeating computation as in CTracker~\cite{peng2020chained}. 
Thus, we can sequentially detect tube boxes for all the frames and generate tube proposals. %Note that the proposals are class-agnostic and the following matching network will classify them to the support classes.

\begin{comment}

\begin{figure}[!t]
\centering
\begin{minipage}[t]{0.45\textwidth}
\includegraphics[width=0.99\textwidth]{figure/TPN.pdf}
\caption{Tube Proposal Network.}
\end{minipage}
\begin{minipage}[t]{0.5\textwidth}
\includegraphics[width=0.99\textwidth]{figure/test-net.pdf}
\caption{Temporal matching (testing).}
\end{minipage}
\end{figure}

\end{comment}

\begin{comment}

\begin{figure}[!t]
\centering
%\fbox{\rule{0pt}{2.5in} \rule{1.0\linewidth}{0pt}}
%0.845
\includegraphics[width=0.5\linewidth]{figure/test-net.pdf}
\vspace{-0.2in}
\caption{{\bf Temporal (tube-based) matching} at inference stage.}
\label{fig:test}
%\vspace{-0.45cm}
\end{figure}

\end{comment}

\vspace{-0.2in}

\subsubsection{Temporal Matching Network}

After obtaining tube proposals, we extract and aggregate tube features and compare them with support features using a matching network, where the matching results are then distributed to the tube proposals in all frames. We re-design the matching network (MN) in the temporal domain to take advantage of tube features. Consequently, our discriminative temporal matching network TMN+ and TPN which share backbone features  are jointly trained for better optimization. Below we detail the design rationale on a single instance/track $i$, starting from MN, TMN and finally TMN+, and it is easy to apply them on multiple objects of different classes.

\noindent{\bf MN.~} 
From $\{I_1, I_2\}$, the \emph{\textbf{query branch}} of backbone extracts query features $\{\mathcal{Q}_1^i, \mathcal{Q}_2^i\}$ for each proposal $p_i$ of instance $i$ with RoIAlign operator. The \emph{\textbf{support branch}} extracts the support features $\mathcal{S}$ in the ground-truth boxes of the support images. The MN then computes the distance between $\mathcal{Q} = \frac{1}{2} (\mathcal{Q}_1^i + \mathcal{Q}_2^i)$ and $\mathcal{S}$ and classifies $\mathcal{Q}$ to the nearest support neighbor. We adopt the multi-relation head with contrastive training strategy from FSOD~\cite{fan2020few} as our matching network (MN) for its high discriminative power. 
Refer to the supplementary material for more details about its architecture.

\noindent{\bf TMN.~} The above MN is however designed for  image object detection and is unsuitable to be applied in the temporal domain. The main problem is the misalignment between  training and inference for the query features $\mathcal{Q}$: In the training stage, $\mathcal{Q}_{train} = \frac{1}{2} (\mathcal{Q}_1^i + \mathcal{Q}_2^i)$ only involves the proposal feature in two frames, limited by the GPU memory and the joint training with TPN. While in the inference stage, $\mathcal{Q}_\text{test}=\frac{1}{T} (\mathcal{Q}_1^i + \mathcal{Q}_2^i + ... + \mathcal{Q}_T^i)$ is derived from all the frames in the tube proposal. This misalignment can produce bad matching result and overall performance degradation. 

To bridge this training and inference gap, we propose a novel temporal matching network (TMN) by introducing a \emph{\textbf{ temporal alignment branch (TAB)}} for query feature alignment. Specifically, %for proposal $p_i$, we first match it with the groundtruth object $i$ in the video which has the largest overlap with the proposal, 
for proposal $p_i$ of the target object $i$,
$\mathcal{Q}_{train} = \frac{1}{2} (\mathcal{Q}_1^i + \mathcal{Q}_2^i)$ involves two frames $\{I_1, I_2\}$,
and the TAB randomly selects images\footnote{
The random selection can be regarded as data augmentation to imitate the imperfect tube features during inference.}
from remaining frames $\{I_3, I_4, ..., I_T\}$ and extracts the aligning features $\mathcal{Q}_{a}=\frac{1}{M} (\mathcal{Q}_3^i + \mathcal{Q}_4^i + ... + \mathcal{Q}_M^i)$ for the target object $i$, where $M$ is the number of selected aligning query images. Then we  generate the aligned query feature $\mathcal{Q}_{ad}^i = \alpha \mathcal{Q}_{train}^i + (1 - \alpha) \mathcal{Q}_{a}^i$ %\footnote{
%\fancomment{The $\alpha$ is used to balance their ratio when using the aligning query feature $\mathcal{Q}_a^i$ to enhance the query feature $\mathcal{Q}_t^i$.}}
as the feature aggregation
to represent the target object and perform matching with supports in the training stage. Our TMN thus bridges this gap %between training and inference
without disrupting the design of TPN and without introducing additional computational overhead by removing TAB at inference time.

The loss function is $\mathcal{L}_{\text{tmn}} = \mathcal{L}_{\text{match}} + \mathcal{L}_{\text{box}}$, where $\mathcal{L}_{\text{match}}$ is the cross-entropy loss for binary matching and $\mathcal{L}_{\text{box}}$ is the smooth $\mathcal{L}_1$ loss for box regression.

\begin{comment}
\begin{figure}[!t]
\centering
%\fbox{\rule{0pt}{2.5in} \rule{1.0\linewidth}{0pt}}
\includegraphics[width=0.95\linewidth]{latex/figure/tpn-2.pdf}
\caption{TPN training (a) and testing (b) stages.
%\textcolor{red}{CK: this figure seems to have {\em never} been referred to in the text!}
}
\label{fig:tpn_illustrate}
\vspace{-0.5cm}
\end{figure}
\end{comment}

\noindent{\bf TMN+.}
To enhance discriminative ability, TMN+ incorporates label-smoothing regularization~\cite{szegedy2016rethinking} into TMN for better generalization and a jointly optimized support classification module for more representative feature. 

We first introduce label smoothing to the matching loss $\mathcal{L}_{\text{match}}$ of TMN, which is widely used to prevent overfitting in the classification task~\cite{muller2019does,thulasidasan2019mixup} by changing the ground-truth label $y_i$ to $y_i^* = (1 - \varepsilon) y_i + \frac{\varepsilon}{\beta}$, where $\varepsilon$ is the constant smoothing parameter and $\beta$ is the number of classes. %\textcolor{red}{(CK: use another letter; $K$ has been used in N-way K-shot.)} \fancomment{
This prevents the model from being overconfident to the training classes and is therefore inherently suitable for the few-shot learning models focusing on the generalization on novel classes. Then, we add a support classification module (classifier) to the support branch to enhance the intra-class compactness and inter-class separability in the Euclidean space and thus generate more representative features for  matching in TMN. We adopt cross-entropy loss as its loss function $\mathcal{L}_{\text{scls}}$.

%\subsection{End-to-End Training}

During training, the TPN and TMN+ are jointly and end-to-end optimized with the weight-shared backbone network by integrating all the aforementioned loss functions:
\begin{equation}
    \mathcal{L} = \lambda_1 \mathcal{L}_{\text{tpn}} + \lambda_2 \mathcal{L}_{\text{tmn}} + \lambda_3 \mathcal{L}_{\text{scls}}
\end{equation}
where $\lambda_1$, $\lambda_2$, and $\lambda_3$ are hyper-parameter weights to balance the loss functions and are set to $1$ in our experiments.

\section{FSVOD-500 Dataset}
\label{sec:dataset}

%\noindent{\bf Existing Video-Level Datasets} 
There exist a number of public datasets with box-level annotations for different video tasks: ImageNet-VID~\cite{deng2009imagenet} for video object detection;
LaSOT~\cite{fan2019lasot}, GOT-10k~\cite{huang2019got}, Youtube-BB~\cite{real2017youtube}, and TrackingNet~\cite{muller2018trackingnet} for single object tracking;  MOT~\cite{milan2016mot16}, TAO~\cite{dave2020tao}, Youtube-VOS~\cite{xu2018youtube} and Youtube-VIS~\cite{yang2019video} for multi-object tracking. 
However, none of these datasets meet the requirement of our proposed few-shot video object detection task. Some datasets (Youtube-BB~\cite{real2017youtube}, TrackingNet~\cite{muller2018trackingnet}, ImageNet-VID~\cite{deng2009imagenet}, Youtube-VOS~\cite{xu2018youtube} and Youtube-VIS~\cite{yang2019video}) have many videos but limited classes, whereas a sufficiently large number of base classes is essential to  few-shot learning. On the other hand, although other datasets (GOT-10k~\cite{huang2019got} and TAO~\cite{fan2019lasot}) contain diverse classes, 
not all instances of the same target class are annotated in a video,
%they are non-exhaustively labeled for target classes in each video (\eg, some instances of target classes are not annotated)}
and therefore are not suitable for the few-shot task.
Last but not least, all of these datasets are not specifically designed for few-shot learning whose train/test/val sets are class-overlapping and cannot be used to evaluate the generality on unseen classes. 

Thus, we design and construct a new dataset for the development and evaluation of few-shot video object detection task. The design criteria are: 
%. There are several standards when collecting the dataset: 
%%\vspace{-0.1in}
\begin{itemize}
    \item The dataset should consist of \emph{\textbf{highly-diversified classes}} for learning a general relation metric for novel classes.
    %%\vspace{-0.05in}
    \item The dataset should be \emph{\textbf{class-balanced}} where each class has similar number of samples to avoid overfitting to any classes, given the long-tailed distribution of many novel classes in the real world~\cite{gupta2019lvis}.%We follow these two criteria to build our dataset. 
    %%\vspace{-0.05in}
    \item The train/test/val sets should contain \emph{\textbf{disjoint}} classes to evaluate the generality of models on novel classes. 
    %%\vspace{-0.1in}
\end{itemize}

\begin{comment}
\begin{figure}[!t]
\centering
%\fbox{\rule{0pt}{2.0in} \rule{1.0\linewidth}{0pt}}
\includegraphics[width=0.8\linewidth]{figure/tree.jpg}
\caption{{\bf FSVOD-500 class hierarchy.} Some third-level node and leaf classes are omitted for simplification. Refer to the supplementary material for the full hierarchy.}
\label{fig:data_fig}
%\vspace{-0.5cm}
\end{figure}
\end{comment}

%\fancomment{To (a good purpose, \eg, take the advantage of the existing datasets, or save labeling labor)}, 
To save human annotation effort as much as possible,
rather than building our dataset from scratch, we exploit existing large-scale video datasets for supervised learning, \ie, LaSOT~\cite{fan2019lasot}, GOT-10k~\cite{huang2019got}, and TAO~\cite{dave2020tao} to construct our dataset subject to the above three criteria.
%Following the above  guidelines, 
The dataset construction pipeline is consist of dataset filtering, balancing and splitting. 
%Following the pipeline of dataset filtering, balancing and splitting, %\footnote{Please refer to the supplementary material for more details of the pipeline.},
%we first construct the train set based on the categories in MS COCO~\cite{lin2014microsoft}
%and then distribute the remaining classes to test/val sets. 
%Specifically, we build a three-levels class tree (shown in the supplementary material) and select third-level node classes similar to the COCO classes and exploit their leaf node as the training classes. The remaining classes are very distinct from COCO classes, and thus used to construct the test/val sets by randomly splitting the node classes into test/val sets. In this way, we can take advantage of the pre-training model on COCO dataset, while the test/val classes are rare novel classes and thus complying to the few-shot setting.

\noindent{\bf Dataset Filtering.~} Note that the above datasets  cannot be  directly used since they are only partially annotated for tracking task: although multiple objects of a given class are present in the video, only some or as few as one of them is annotated while the others are not annotated. Thus, we filter out videos with non-exhaustive labels while keeping those with high-quality labels covering all objects in the same class (target class). We also remove  videos containing extremely small objects which are usually in bad visual quality and thus unsuitable for few-shot learning. 
Note that exhaustive annotation for all possible classes in such a large dataset is expensive and infeasible~\cite{dave2020tao,gupta2019lvis}. Therefore, only the target classes are exhaustively annotated for each video while non-target classes are categorically ignored.

%\vspace{1pt}

\noindent{\bf Dataset Balancing.~} It is essential to maintain good data balancing in the few-shot learning dataset, so that sufficient  generality to novel classes can be achieved without overfitting to any dominating training classes. Thus, we  remove  `person' and `human face' from the dataset which are in  massive quantities (and they have already been extensively studied in many works and tasks~\cite{ess2008mobile,dollar2009pedestrian,xiao2017joint,kemelmacher2016megaface}).%,zhang2017citypersons,wolf2011face}). 
Then, we manually remove easy samples for those classes with more than 30 samples. Finally, each class in our dataset has at least 3 videos and no more than 30.

%\vspace{1pt}

\noindent{\bf Dataset Splitting.}
We summarize a four-level label system (shown in the supplementary material) to merge these datasets by grouping their leaf labels with the same semantics (\eg, truck and lorry) into one class. Then, we select third-level node classes similar to the COCO~\cite{lin2014microsoft} classes and exploit their leaf node as the training classes. The remaining classes are very distinct from COCO classes, and thus used to construct the test/val sets by randomly splitting the node classes. In this way, we can take advantage of the pre-training model on COCO dataset, while the test/val classes are rare novel classes and thus complying to the few-shot setting.
We follow three guidelines for dataset split:
\\
{\bf G1:} The split should be in line with the few-shot learning setting, \ie, train set should contain common classes in the real world, while  test/val sets should contain rare classes.
{\bf G2:} To take advantage of pre-training on other datasets,  %\fancomment{(old: for better performance. Maybe we need a better reason)}, 
the train set should have a large overlap with existing datasets while the test/val sets should have largely no overlap.
{\bf G3:} The train and test/val sets should have different node classes  to evaluate the generality on novel classes in a challenging setting to avoid the influence of similar classes across sets, \eg, if the train set has `Golden Retriever', it is much easier to detect `Labrador Retriever' in the test set, which is undesirable.

Consequently,  \emph{\textbf{FSVOD-500}} is the first benchmark specially designed for few-shot video object detection in evaluating the performance of 
a given model on novel classes.% in the video domain.

%\noindent{\bf Dataset Statistics.~} Our   {\bf FSVOD-500} dataset is the first benchmark specifically designed for few-shot video object detection for evaluating the generality of a proposed model on novel classes in  the video domain, containing 500 classes with 320/80/100 classes for training/val/test,  with at least 3 videos and no more than 30 videos in each class. There are respectively xxx/xxx/xxx videos derived from GOT-10k/LaSOT/TAO dataset, with around xxxxx images and xxxxxx bounding boxes in total. Each video has been meticulously and exhaustively annotated at 1 FPS and contains at least one object.

\begin{table}[!t]
\begin{center}
\tabcolsep=12pt 
\caption{{\bf Dataset statistics} of FSVOD-500 and FSYTV-40. ``Class Overlap'' denotes the class overlap with MS COCO~\cite{lin2014microsoft} dataset.}
%\normalsize
\vspace{-0.1in}
\begin{tabular}{lccccc}
 & \multicolumn{3}{c}{{\bf FSVOD-500}} & \multicolumn{2}{c}{{\bf FSYTV-40}} \\
 & Train & Val & Test & Train & Test \\
\hline

\hline
%\hline
\rowcolor{Gray1}label FPS & 1 & 1 & 1 & 6 & 6 \\
\rowcolor{Gray2} \# Class & 320 & 80 & 100 & 30 & 10 \\
\rowcolor{Gray1}\# Video & 2553 & 770 & 949 & 1627 & 608 \\
\rowcolor{Gray2} \# Track & 2848 & 793 & 1022 & 2777 & 902 \\
\rowcolor{Gray1}\# Frame & 60432 & 14422 & 21755 & 41986 & 19843 \\
\rowcolor{Gray2} \# Box & 65462 & 15031 & 24002 & 66601 & 27924 \\
\hline
\rowcolor{Gray1}Class Overlap & Yes & No & No & Yes & No \\
\rowcolor{Gray2}Exhaustive & \multicolumn{3}{c}{Only target classes} & \multicolumn{2}{c}{All classes} \\
%Avg No. Box / Img  & 2.82 & 2.48 & \\ %per Image
%Min No. Img / Cls  & xxx & xxx & xxx \\ %per Class
%Max No. Img / Cls  & xxx & xxx & xxx \\ %per Class
%Avg No. Img / Cls  & xxx & xxx & xxx \\ %per Class
%Box Size & [6, 6828] & [13, 4605] &  \\
%Box Area Ratio & [0.0009, 1] & [0.0009, 1] &  \\
%Box W/H Ratio & [0.0216, 89] & [0.0199, 51.5] &  \\
\end{tabular}

\end{center}
%\vspace{-0.1in}

\label{table:data}
\vspace{-0.3in}
\end{table}

\section{Experiments}

We conduct extensive experiments to validate the effectiveness of our proposed approach. Since this is the first paper on FSVOD, we compare with state-of-the-art (SOTA) methods of related tasks by adapting them to the FSVOD task. %on  FSVOD-500. %and  FSYTV-40.

%\subsection{Experiment Details}

\noindent{\bf Training.~} Our model is trained on four GeForce GTX 1080Ti GPUs using the SGD optimizer with 45,000 iterations. The initial learning rate is set to 0.002 which decays by a factor of 10 respectively in 30,000 and 40,000 iterations. Each GPU contains five cropped support images, two query images and $M$ cropped aligning query images in the same video, where $M$ is randomly sampled from $[1,10]$. We use ResNet50~\cite{he2016deep} as our backbone which is pre-trained on ImageNet~\cite{deng2009imagenet} and MS COCO~\cite{lin2014microsoft}\footnote{There is no overlap between MS COCO and the val/test sets of both FSVOD-500 and FSYTV-40 datasets.} for stable low-level features extraction and better convergence. The model is trained with 2-way 5-shot contrastive training strategy proposed in FSOD~\cite{fan2020few}. Other hyper-parameters are set as $\alpha = 0.5, \varepsilon=0.2, \beta=2$ in our experiments.

%The stride of the Res5 block is reduced to 1 to increase feature map resolution and we replace its regular convolution with the dilated convolution to keep the effective receptive field. The low-level layers (Res1 and Res2) are fixed and we only train the high-level layers to utilize the low-level features from the pre-trained model and prevent over-fitting. 
%As for the inputs, the query images are resized such that the shorter size has 600 pixels and longer size is capped at 1000 pixels. We also use multi-scale training for the query images. Both the support images are aligning query images are cropped around the target object with 16-pixel image context, zero-padded and resized  to a square image of $320 \times 320$. For evaluation, we adopt the 5-shot full-way evaluation in our experiments with multiple metrics to evaluate models from different aspects, \ie, \emph{AP}, \emph{AP$_{50}$}, \emph{AP$_{75}$} for  object detection performance, ${\mathit MOTA}$~\cite{} for  tracking performance, ${\mathit Accuracy}$ to evaluate the matching accuracy, and ${\mathit Recall}_{50}$ for recall performance.

\noindent{\bf Evaluation.~} We adopt the full-way 5-shot evaluation 
(exploit all classes in the test/val set with 5 images per class as supports for evaluation) 
in our experiments with standard object detection evaluation metrics, \ie, \emph{AP}, \emph{AP$_{50}$}, and \emph{AP$_{75}$}.
%We also adopt other metrics to evaluate models from different aspects, \ie,  ${\mathit MOTA}$ for  tracking performance, ${\mathit Recall}$ for recall performance, and ${\mathit Accuracy}$ for matching accuracy. 
%\fancomment{
The evaluations are conducted 5 times on randomly sampled support sets and the mean and standard deviation are reported. Refer to the supplemental material for more training and evaluation details. 

\noindent{\bf FSYTV-40.~} To validate model generalization  on datasets with different characteristics, we construct another dataset built on  Youtube-VIS dataset~\cite{yang2019video} for the FSVOD task.  FSYTV-40 is vastly  different from  FSVOD-500 with only 40 classes (30/10 train/test class split following the same dataset split guidelines above, 
with instances of all classes are exhaustively annotated in each video), more videos in each class and more objects in each video. %This dataset contains xxxx videos annotated at 6 FPS, xxxx images and xxxx bounding boxes in total. 
Table~\ref{table:data} tabulates the 
detailed statistics of both datasets.  

\subsection{Comparison with Other Methods}

With no recognized previous work on FSVOD, we adapt representative models from related tasks to perform FSVOD, such as image object detection (Faster R-CNN~\cite{ren2015faster}, %TFA~\cite{wang2020frustratingly} 
and FSOD~\cite{fan2020few}), video object detection (MEGA~\cite{chen2020memory} and RDN~\cite{deng2019relation}) and multiple object tracking (CTracker~\cite{peng2020chained}, and FairMOT~\cite{zhan2020simple}, and CenterTrack~\cite{zhou2020tracking}). Only FSOD model can be directly applied frame-by-frame to perform FSVOD. %We use the imprinting technique~\cite{qi2018low} to generate  classifier weights for TFA.
%, but other models can not directly applied on the FSVOD task. 
For others, we exploit their models to generate class-agnostic boxes and adopt the multi-relation head trained in the FSOD~\cite{fan2020few} model to evaluate the distance between the query boxes and supports.  
%For others, we use the multi-relation head~\cite{fan2020few} to replace the detectors of these models and jointly train them for a fair comparisons.
We first perform comparison on FSVOD-500, and then generalize to FSYTV-40  (Table~\ref{table:maintable}).

\noindent{\bf Comparison with IOD-based methods.~} 
FSOD serves as a strong baseline with its high recall of attention-RPN and powerful generalization of multi-relation head. With the same matching network, Faster R-CNN produces inferior performance due to the lower recall of its generated boxes. %TFA obtains comparable performance to FSOD.
With the representative aggregated query feature from TPN and discriminative TMN+ in the temporal domain, 
our FSVOD model outperforms FSOD by a large margin.

\noindent{\bf Comparison with VOD-based methods.~}
VOD-based methods operate similarly to IOD-based methods in its per-frame object detection %\footnote{Both MEGA and RDN do not produce trajectories.}
followed by matching with supports and thus both suffer from noisy proposals and less powerful features. 
%The VOD-based methods performs better than the IOD-based methods because of better detection results. %from the joint training.
Interestingly, we find that VOD-based methods have a worse performance %than IOD-based methods because the former
because they produce excessive proposals which heavily burden the subsequent matching procedure despite their higher recalls.

\begin{table}[!t]

\begin{center}
\tabcolsep=2pt 
\caption{{\bf Experimental results on FSVOD-500 and FSYTV-40 test set} for novel classes with the full-way 5-shot evaluation.}
\label{table:maintable}

\vspace{-0.1in}
\begin{tabular}{cccccccc}
%\hline

%\hline
& & \multicolumn{3}{c}{{\bf FSVOD-500}} & \multicolumn{3}{c}{{\bf FSYTV-40}} \\
Method & Tube & $AP$ & $AP_{50}$ & $AP_{75}$ & $AP$ & $AP_{50}$ & $AP_{75}$ \\
\hline

\hline
\rowcolor{Gray1}FR-CNN~\cite{ren2015faster} & \xmark      & 18.2$_{\pm 0.4}$ & 26.4$_{\pm 0.4}$ & 19.6$_{\pm 0.5}$   & 9.3$_{\pm 1.4}$ & 15.4$_{\pm 1.7}$ & 9.6$_{\pm 1.7}$  \\
%\rowcolor{Gray2}TFA~\cite{wang2020frustratingly} & \xmark  & 20.5$_{\pm 0.4}$ & 31.0$_{\pm 0.8}$ & 21.5$_{\pm 0.5}$ & 12.0$_{\pm 1.2}$ & 20.8$_{\pm 1.6}$ & 12.7$_{\pm 1.3}$  \\
\rowcolor{Gray2}FSOD~\cite{fan2020few} & \xmark  & 21.1$_{\pm 0.6}$ & 31.3$_{\pm 0.5}$ & 22.6$_{\pm 0.7}$ & 12.5$_{\pm 1.4}$ & 20.9$_{\pm 1.8}$ & 13.0$_{\pm 1.5}$   \\
\hline
\rowcolor{Gray1}MEGA~\cite{chen2020memory} & \xmark & 16.8$_{\pm 0.3}$ & 26.4$_{\pm 0.5}$ & 17.7$_{\pm 0.3}$ & 7.8$_{\pm 1.1}$ & 13.0$_{\pm 1.9}$ & 8.3$_{\pm 1.1}$ \\
\rowcolor{Gray2}RDN~\cite{deng2019relation} & \xmark & 18.2$_{\pm 0.4}$ & 27.9$_{\pm 0.4}$ & 19.7$_{\pm 0.5}$ & 8.1$_{\pm 1.1}$ & 13.4$_{\pm 2.0}$ & 8.6$_{\pm 1.1}$ \\
%\rowcolor{Gray1}MEGA~\cite{chen2020memory} & \xmark & 21.8$_{\pm 0.5}$ & 31.9$_{\pm 0.6}$ & 23.1$_{\pm 0.6}$ & 12.9$_{\pm 1.0}$ & 21.3$_{\pm 1.7}$ & 13.4$_{\pm 1.3}$  \\
%\rowcolor{Gray2}RDN~\cite{deng2019relation} & \xmark & 21.6$_{\pm 0.4}$ & 31.7$_{\pm 0.6}$ & 23.2$_{\pm 0.7}$ & 12.8$_{\pm 0.9}$ & 21.4$_{\pm 1.8}$ & 13.3$_{\pm 1.3}$   \\
\hline
\rowcolor{Gray1}CTracker~\cite{peng2020chained} & \cmark  & 20.1$_{\pm 0.4}$ & 30.6$_{\pm 0.7}$ & 21.0$_{\pm 0.8}$ & 8.9$_{\pm 1.4}$ & 14.4$_{\pm 2.5}$ & 9.1$_{\pm 1.3}$  \\
\rowcolor{Gray2}FairMOT~\cite{zhan2020simple} & \cmark  & 20.3$_{\pm 0.6}$ & 31.0$_{\pm 1.0}$ & 21.2$_{\pm 0.8}$ & 9.6$_{\pm 1.6}$ & 16.0$_{\pm 2.2}$ & 9.5$_{\pm 1.4}$  \\
\rowcolor{Gray1}CenterTrack~\cite{zhou2020tracking} & \cmark & 20.6$_{\pm 0.4}$ & 30.5$_{\pm 0.9}$ & 21.9$_{\pm 0.4}$ & 9.5$_{\pm 1.6}$ & 15.6$_{\pm 2.0}$ & 9.7$_{\pm 1.3}$ \\
%\rowcolor{Gray1}CTracker~\cite{peng2020chained} & \cmark  & 22.5$_{\pm 0.3}$ & 33.2$_{\pm 0.5}$ & 23.6$_{\pm 0.6}$ & 13.3$_{\pm 1.5}$ & 21.7$_{\pm 2.3}$ & 13.8$_{\pm 1.2}$  \\
%\rowcolor{Gray2}FairMOT~\cite{zhan2020simple} & \cmark  & 23.1$_{\pm 0.6}$ & 33.4$_{\pm 0.8}$ & 24.3$_{\pm 0.6}$ & 13.5$_{\pm 1.4}$ & 21.9$_{\pm 2.4}$ & 14.1$_{\pm 1.2}$  \\
%\rowcolor{Gray1}CenterTrack~\cite{zhou2020tracking} & \cmark & 22.7$_{\pm 0.5}$ & 33.1$_{\pm 0.7}$ & 23.9$_{\pm 0.6}$ & 13.2$_{\pm 1.3}$ & 21.6$_{\pm 1.9}$ & 13.7$_{\pm 1.2}$ \\
\hline
\rowcolor{Gray2}Ours & \cmark & \textbf{25.1$_{\pm 0.4}$} & \textbf{36.8$_{\pm 0.5}$} & \textbf{26.2$_{\pm 0.7}$} & \textbf{14.6$_{\pm 1.6}$} & \textbf{21.9$_{\pm 2.0}$} & \textbf{16.1$_{\pm 2.1}$}   \\   
%\rowcolor{Gray2}Ours & \cmark & \textbf{26.2$_{\pm 0.5}$} & \textbf{38.7$_{\pm 0.7}$} & \textbf{27.6$_{\pm 0.6}$} & \textbf{15.2$_{\pm 1.8}$} & \textbf{22.7$_{\pm 2.2}$} & \textbf{16.6$_{\pm 2.3}$}   \\          
%\rowcolor{Gray2}Ours$^\ddagger$ & & & & \textbf{18.3$_{\pm 0.7}$} & \textbf{28.4$_{\pm 1.2}$} & \textbf{19.5$_{\pm 0.8}$}  \\
%\hline

%\hline
\end{tabular}
\end{center}
\vspace{-0.3in}
\end{table}

\noindent{\bf Comparison with MOT-based methods.~}
MOT-based methods have a similar detection mechanism to our approach, by first generating tubes for query objects and representing them with the aggregated tube features, followed by matching between query tube features and support features. Thus,
%\fancomment{
even with much lower recalls ($\sim$$70.0 \%$ \textit{v.s.} $\sim$$80.0 \%$), they still have better performance %\footnote{We lower the detection threshold of MOT-based methods for higher recall and better AP at the expense of tracking performance.}
than VOD-based methods by taking advantage of temporal matching.
%} 
However, our approach still outperforms MOT-based methods by a significant margin leveraging our jointly optimized TPN and TMN+ with more representative features and powerful matching network.

\noindent{\bf Generalization on FSYTV-40 dataset.~}
This dataset is very different from FSVOD-500 with the former having significantly less classes but more videos in each class, more tracks in each video, and higher annotation FPS. Although our method still outperforms other methods on this dataset, a substantial performance degradation in comparison with  FSVOD-500 is resulted, which is caused by the much reduced  class diversity for the matching network to learn a general relation metric for novel classes. To verify this, we train our model on the FSVOD-500 train set and evaluate it on the FSYTV-40 test set\footnote{There is no overlapping or similar classes between them.}. 
It can promote the performance from 14.6 to 17.8 AP.
The resulting large performance boost again validates  the importance of high diversity of training classes, one of the desirable properties of % and the effectiveness of 
our FSVOD-500 for few-shot video object learning.

\subsection{Ablation Studies}

%\fancomment{
Table~\ref{table:ablation} tabulates the ablation studies on the proposal box generation network and matching (classification) network. Compared to RPN, our proposed TPN improves the performance by 3.7 AP with the same matching network. Although RPN and TPN have similar recall performance ($76.2 \%$ vs $76.8 \%$), TPN has a better classification performance due to its discriminative and  aggregated temporal features, and therefore producing better detection and matching performance.
%}

%\fancomment{
For the matching network, the RN (Relation Network~\cite{yang2018learning}) based baseline performs worst which is limited by its weak matching ability. Replacing RN by the more powerful multi-relation MN~\cite{fan2020few} can significantly improve the performance. 
When cooperating with  TPN,
our proposed TMN outperforms MN by 3.2 AP in the temporal domain using aligned query features.
The improved  TMN+ reaches 30.0 AP performance by capitalizing on better generalization and representative feature, which is optimized with the label-smoothing regularization and support classification module, bringing about respectively $1.5$ and $3.0$ performance increase. %($26.4+1.3+3.0=30.7$). 
%Note that for MN and TMN, their improved versions MN++ and TMN+ respectively outperform the original MN and TMN by capitalizing on better generalization and representative feature optimized with the label-smoothing regularization and support classification module,  bringing about respectively $00.0$ and $00.0$ increase in performance.
%}
%\fancomment{
Note that our support classification module is fundamentally different from the meta-loss in Meta R-CNN~\cite{yan2019metarcnn} which requires training on novel classes to avoid prediction ambiguity in object attentive vectors, while our method targets at generating more representative features in the Euclidean space to generalize better on novel classes without any fine-tuning. 
%Because of feature frames reuse, the inference is efficient at around 184 ms per image. %and the model computation complexity is 174.5 GFLOPS.}

%}

\begin{table}[!t]
\begin{center}

\tabcolsep=8pt 
\caption{{\bf Ablation experimental results on FSVOD-500 val set} for 80 novel classes with the full-way 5-shot evaluation. ``LSR'' denotes  label-smoothing regularization and ``SCM'' denotes  support classification module.}
\label{table:ablation}
\vspace{-0.1in}

\begin{tabular}{ccccc}

Box & Matching & $AP$ & $AP_{50}$ & $AP_{75}$ \\
\hline

\hline
\multirow{ 2}{*}{RPN} & \cellcolor{Gray1} RN    & \cellcolor{Gray1} 10.1$_{\pm 0.5}$ & \cellcolor{Gray1} 14.0$_{\pm 0.6}$ & \cellcolor{Gray1} 11.1$_{\pm 0.7}$  \\
                      & \cellcolor{Gray2} MN    & \cellcolor{Gray2} 19.5$_{\pm 0.9}$ & \cellcolor{Gray2} 27.4$_{\pm 1.2}$ & \cellcolor{Gray2} 21.8$_{\pm 1.1}$  \\
                      %& MN++  & 25.4$_{\pm 0.7}$ & 35.4$_{\pm 1.0}$ & 28.1$_{\pm 0.7}$  \\
\hline
\multirow{ 5}{*}{TPN} & \cellcolor{Gray1} MN    & \cellcolor{Gray1}23.2$_{\pm 1.2}$ & \cellcolor{Gray1}32.7$_{\pm 1.5}$ & \cellcolor{Gray1}25.6$_{\pm 1.5}$  \\
                      %& MN++  & 28.0$_{\pm 0.4}$ & 39.6$_{\pm 0.7}$ & 30.8$_{\pm 0.5}$  \\
                      & \cellcolor{Gray2}TMN   & \cellcolor{Gray2}26.4$_{\pm 1.5}$ & \cellcolor{Gray2}37.2$_{\pm 1.4}$ & \cellcolor{Gray2}29.5$_{\pm 1.6}$  \\
                      & \cellcolor{Gray1} TMN w/ LSR & \cellcolor{Gray1}27.9$_{\pm 1.3}$ & \cellcolor{Gray1}39.6$_{\pm 1.2}$ & \cellcolor{Gray1}30.8$_{\pm 1.5}$  \\
                      & \cellcolor{Gray2}TMN w/ SCM & \cellcolor{Gray2}29.4$_{\pm 0.8}$ & \cellcolor{Gray2}41.8$_{\pm 1.1}$ & \cellcolor{Gray2}31.9$_{\pm 1.2}$  \\ 
                      & \cellcolor{Gray1} TMN+ & \cellcolor{Gray1}\textbf{30.0$_{\pm 0.8}$} & \cellcolor{Gray1}\textbf{43.6$_{\pm 1.2}$} & \cellcolor{Gray1}\textbf{32.9$_{\pm 1.1}$}  \\ 
\end{tabular}
\end{center}
%\vspace{-0.1in}
\vspace{-0.35in}
\end{table}

\subsection{Advantages of Temporal Matching}
Temporal matching %is the key idea of our approach which 
has two substantial advantages over image-based matching:

\noindent{\bf Ghost Proposal Removal.~} 
Image-based matching suffers heavily from  ``ghost proposals" which are  hard background proposals with similar appearance to foreground proposals. It is difficult to filter them out by the RPN in the spatial domain due to  appearance ambiguity, while much easier to distinguish in the temporal domain due to their intermittent ``ghost" or discontinuous appearances  across frames. Our TPN  takes this advantage to get rid of ghost proposals and thus obtains better detection performance. 

\noindent{\bf Representative Feature.~} 
From the feature perspective, image-based matching exploits proposal features from each query frame to match with supports individually. 
\begin{comment}
{\bf (CK: check above sentence. Qi: Checked. \fancomment{Qi: The key is that the query proposal feature in image-based matching is only from one image. Maybe we should emphasize the unstable independent query feature)}} 
\end{comment}
Such independent query feature is inadequate in representing a target video object, especially those in bad visual quality due to \eg, large deformation, motion blur or heavy occlusion, thus is liable to bad comparison results in the subsequent matching procedure and leading to bad predictions. In contrast, our temporal matching aggregates object features across frames in the tube proposal into a robust representative feature for the target video object, which helps the subsequent matching procedure to produce better result. %matching result.

\begin{figure*}[!t]
\centering
%\fbox{\rule{0pt}{2.5in} \rule{0.99\linewidth}{0pt}}
\includegraphics[width=0.8\linewidth]{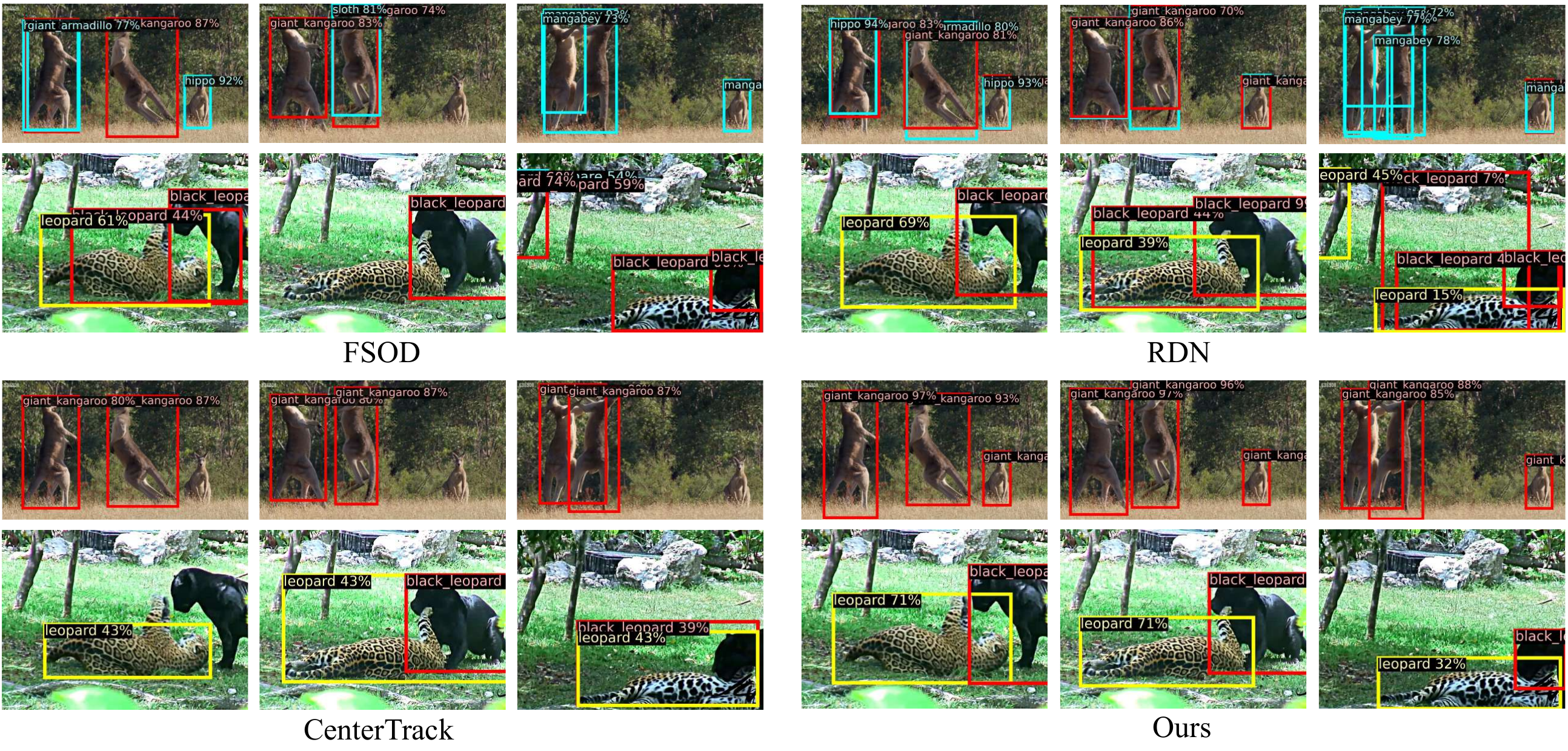}
\vspace{-0.2in}
\caption{
{\bf Qualitative 5-shot detection results on novel classes of FSVOD dataset}. %(supports are omitted for clarity). 
Our tube-based approach successfully detects objects in novel classes, while other methods miss or misclassify target objects or detect ghost objects. %The ground-truth class labels of each video are ``corvette'' and ``guard ship'' (top), ``giant kangaroo'' (middle), and ``leopard'' and ``black leopard'' (bottom) respectively.
}
\label{fig:vis}
\vspace{-0.15in}
\end{figure*}

\noindent{\bf Validation.~} We show quantitatively and qualitatively the above advantages of our temporal matching. 
%\fancomment{
Specifically, we transform our tube-based matching to the image-based matching by performing per-frame detection and matching during inference. With the same trained model, the performance drastically drops from 30.0 to 25.8 after replacing  tube-based feature by image-based feature.
%}
The large performance gap indicates the effectiveness of tube-based matching in the FSVOD task. In Figure~\ref{fig:vis}, the image-based methods produce ghost proposals and fails the target object matching, %\fancomment{(Qi: there are two error types corresponding to the ghost proposal and worse feature)}
while our approach produces much better performance without suffering from ghost proposals.

\subsection{Object Indexing in Massive Videos}

Our FSVOD task enables models properly solving the object indexing/retrieval problem in massive videos, which is infeasible or extreme hard for other computer vision tasks.
Specifically, we retrieve video clips for the target support class if there exists a detected box with the class score larger than 0.05.
Thanks to the full-way evaluation, our FSVOD actually performs indexing for every class in the entire video set.
We use the widely-used $F_1$ score to evaluate the retrieval performance. Our FSVOD model achieves $0.414$ $F_1$ score on FSVOD-500 test set, while the classic few-shot object detection model~\cite{fan2020few} only obtains $0.339$ $F_1$ score because of its numerous false positive predictions in videos. 
More details are in the supplementary material.

\vspace{-0.1in}
\section{Conclusion}
\vspace{-0.05in}

This paper proposes
 FSVOD for detecting objects in novel classes in a query {\em video}
 given {\em only a few support images}.
 FSVOD can be applied in high diversity/dynamic scenarios for solving relevant real-world problem that is infeasible or hard for other computer vision tasks.
 We contribute a new large-scale, class-balanced FSVOD dataset, which contains 500 classes of objects in high diversity with high-quality annotations. Our tube proposal network and aligned matching network effectively employ the temporal information in  proposal generation and matching. Extensive comparison have been performed to compare related methods on two datasets to validate that our FSVOD method produces the best performance. 
 We hope this paper will kindle future FSVOD research.%, an important problem in computer vision.

\section{More Implementation Details}

\subsection{Matching Network Architecture}

We adopt the multi-relation head from FSOD~\cite{fan2020few} as our matching network (MN), which consists of three relation heads for learning to match support and query features $\{f_s, f_q\} \in \mathbb{R}^{1 \times C \times 7 \times 7}$ in multiple levels. 

{\bf Global-relation head.} Designed to learn a global matching embedding, this head first concatenates $f_s$ and $f_q$ along the channel dimension to feature $f_c^{'} \in \mathbb{R}^{1 \times 2C \times 7 \times 7}$, which is  then average pooled  to $f_c \in \mathbb{R}^{1 \times 2C \times 1 \times 1}$. Finally, a MLP $\mathcal{M}$ containing three fully connected layers with ReLU (except the last one) is applied to $f_c$ to predict the matching score $s_g = \mathcal{M}(f_c)$. 

{\bf Patch-relation head.} Designed to learn a non-linear metric to capture the complex relation between patches, this head is derived from the RelationNet~\cite{sung2018learning} where the concatenated feature $f_c^{'}$ is fed to a small convolution network, which  consists of two $3 \times 3$ average pooling operators at the first and last layers separately, two $1 \times 1$ convolutional layers for reducing and then restoring dimensions, and one $3 \times 3$ convolutional layer (all convolutional layers are equipped with ReLU). Note that all these operations and layers use one stride and zero padding to generate the final feature vector $f_q \in \mathbb{R}^{1 \times C \times 1 \times 1}$. Finally, a fully connected layer is employed to generate the matching score $s_p$, and a sibling fc layer to generate the box prediction for better supervision from multi-task learning.

{\bf Local-relation head.} Designed to capture the pixel-level relation between support and query features,
$f_s$ and $f_q$ are first  processed using a weight-shared convolution layer with ReLU. Then their pixel-wise relation is calculated by depth-wise correlation~\cite{li2019siamrpn++} with the resulting feature vector $f_d \in \mathbb{R}^{1 \times C \times 1 \times 1}$  fed to a fully connected layer to generate the matching score $s_l$.

These three relation heads cooperate together to capture the relation between support and query features in different levels. The final matching score is obtained by summing all the aforementioned matching scores: $s = s_g + s_p + s_l$.

\subsection{Deformable RoIAlign}

Deformable RoIAlign dynamically changes its sample locations according to the input features. In our implementation, two frame features are concatenated and sent to the deformable RoIAlign so that it is aware of the object positions in both frames and therefore dynamically adapt the sample locations to enlarge the search region to capture objects in both frames.

\subsection{Training Details}

The stride of the Res5 block is reduced to 1 to increase feature map resolution. We  replace its regular convolutional layer with the dilated convolutional layer to keep the effective receptive field.
Following common practices, the low-level layers (Res1 and Res2) are fixed and only the high-level layers are trained.
%The low-level layers (Res1 and Res2) are fixed and we only train the high-level layers to utilize the low-level features from the pre-trained model and prevent over-fitting. 
As for the inputs, the query image is resized to (600, 1000) where the shorter and longer sizes are respectively no longer than 600 and 1000 pixels. We also adopt the multi-scale training for query images during training. As for the support and aligning query images, they are cropped and resized to $320 \times 320$ size with extended 16-pixels around the target object and the cropped images are saved to the disk for efficient training to avoid repeating the crop for the same image.
%are resized such that the shorter size has 600 pixels and longer size is capped at 1000 pixels. We also use multi-scale training for the query images. 
%Both the support and aligning query images 

%Both the support images are aligning query images are cropped around the target object with 16-pixel image context, zero-padded and resized to a square image of $320 \times 320$.

\subsection{Evaluation Details}

During inference, the final score of each box is obtained by multiplying the matching score predicted by TMN and the corresponding objectness score generated by TPN % as the final score for each box. 
%It enables the model 
to suppress the scores of boxes containing hard background\footnote{The background may have high matching score because of the similar appearance with supports. The low objectness score predicted by TPN can down-weigh the overall score to alleviate this influence.}. %and slightly improves the performance.

Instead of setting a fixed support set which is only used for support images, we exploit a support set which can fully utilize the val and test set in a dynamic manner for more comprehensive evaluation on all videos. %(denoted as ``dynamic support set'').

The following description applies to the val set which is similar to the  test set. Our {\em dynamic} support set contains ``offline'' and ``online'' support sets. The support images in the offline support set are derived from the randomly selected val set videos $V_{\text{offline}}$. The support features $f_{\text{offline}}$ are pre-computed and saved to the hard drive for efficient evaluation\footnote{One class has one corresponding support feature in the $C \times 1 \times 1$ size and $C$ is the channel number.}. Then we can load the pre-computed support features to the model to perform detection on query videos. When performing evaluation on $V_{\text{offline}}$, we build the online support sets by randomly selecting images from the remaining videos, and the support features are online generated for the evaluation\footnote{We only use them for $V_{\text{offline}}$ with a small number of videos without reusing again. It is also feasible to first save them to the hard drive.}. In this way, we avoid the ``cheap matching'' between same objects which is degraded to the single object tracking task. The dynamic support set can dynamically decide the support sets for different videos, and therefore efficiently utilizes the entire val set to perform evaluation without leaving a fraction of videos as the specialized support set. Note that the video-level annotation is much more expensive and time-consuming than the image-level annotation. With our dynamic support set we can avoid wastage of valuable video data.

Note that the finetuning-based methods cannot be directly compared with matching-based methods because of the former's high requirement for support sets, which requires training on novel classes in a reserved support set. This limits the application of the finetuning-based methods, because it is impossible to exhaustively annotate all  videos\footnote{In our case, for a novel class, we need to annotate at least one video and one support image containing a different object belonging to the same class to avoid ``cheap matching''.} for each novel class. This issue can be solved by annotating a special support set (it is very time and money consuming) to finetune these models. 
% don't say the following
% discuss that in future when time comes
%However, since it is deviated from the discussion of our paper and the annotation task is demanding, we will discuss it in the future.

%The only limitation is that since there is no reserved support set ,  finetuning-based methods cannot be trained on the val/test classes and therefore cannot be directly compared with our method on the entire val/test set. Also, the high requirement for support sets limits the application of the finetuning-based methods because it is impossible to annotate videos\footnote{For a novel class, we need to annotate at least one video and one support image containing a different object belonging to the same class to avoid ``cheap matching''.} for each novel class. We will build another test set in the future to fairly compare the matching-based and finetuning-based methods.

%first set ``offline'' and ``online'' support sets where the offline support set is fixed and the support features of this set are pre-computed 

\section{Error Type Analysis}

%\fancomment{
To conduct an in-depth investigation of different models on the FSVOD task, we analyze the error types on four representative models, namely, FSOD~\cite{fan2020few}, RDN~\cite{deng2019relation}, CenterTrack~\cite{zhou2020tracking} and our FSVOD using a general toolbox TIDE~\cite{bolya2020tide} which segments object detection errors into six types and measures the contribution of each error by isolating its effect on overall performance (refer to~\cite{bolya2020tide} for more details).

%by isolating its effect on overall performance as in TIDE~\cite{bolya2020tide} (refer to~\cite{bolya2020tide} for more details). Specially, the 

%, using a general toolbox TIDE~\cite{bolya2020tide} which segments object detection errors into six types and measures the contribution of each error by isolating its effect on overall performance (refer to~\cite{bolya2020tide} for more details).
%}

\begin{figure}[!t]
\centering
%\fbox{\rule{0pt}{2.5in} \rule{1.0\linewidth}{0pt}}
\includegraphics[width=0.6\linewidth]{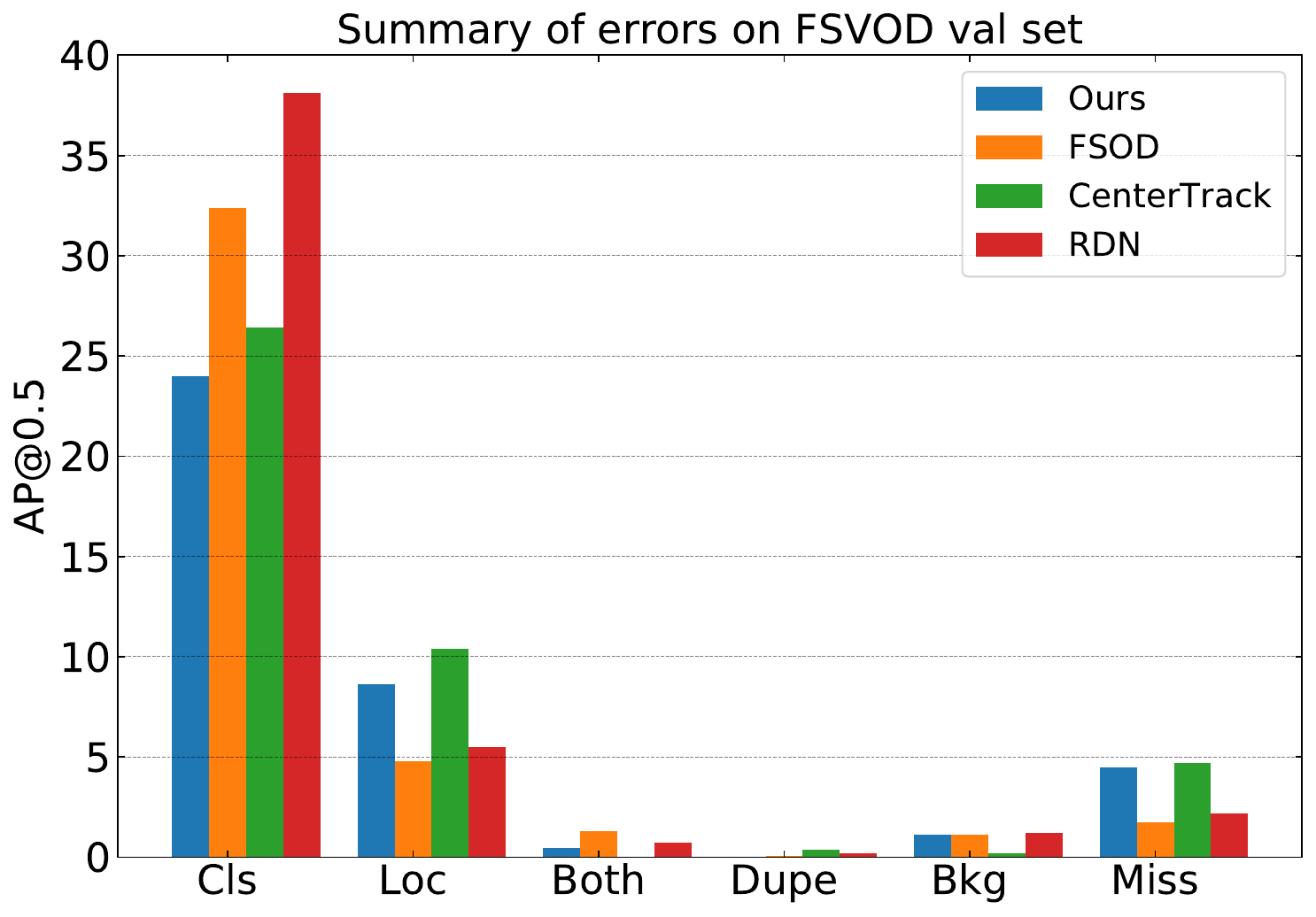}
\caption{Error type analysis of different methods on FSVOD-500 val set. Lower is better.}
\label{fig:tide}
\vspace{-0.2cm}
\end{figure}

%\fancomment{
Figure~\ref{fig:tide} indicates that all methods suffer from  classification errors on our FSVOD-500 dataset, revealing that the core problem of FSVOD lies on few-shot learning in distinguishing and classifying novel classes, which cannot be replaced by other video understanding tasks. 
%This substantially differentiates from other video understanding tasks and this distinctive characteristic further confirms the importance and necessity of our FSVOD task.
%}

%\fancomment{
For individual performances: FSOD has the lowest localization and missing error thanks to the high-quality proposals generated by its attention RPN;
%FSOD differentiates from other methods in its much higher false positive error and low localization related errors deriving from  massive object  proposals 
%(Qi comment: I want to express that most of these proposals covering objects because of the attention RPN, so that it can locate all the objects and tends to misclassify other objects) 
%generated by its attention RPN; 
the VID-based model RDN mainly suffers from  classification error %with lower localization related errors 
because it generates too many background  proposals which exacerbate the following matching procedure; the MOT-based model CenterTrack has lower classification error benefiting from the robust tube-based feature, but it suffers higher localization and missing errors caused by its lower recall. 
Our approach has the lowest classification error benefiting from our strategically designed TMN+ which leverages the representative tube-based features generated by TPN.
%} 

%\fancomment{
From the above error analysis, we  conclude that solving the few-shot matching problem is the most essential future direction for FSVOD.
We show more experimental results under different few-shot evaluation settings in Table~\ref{table:setting}.
%}

\begin{table}[!t]
\begin{center}
\tabcolsep=8pt 

\begin{tabular}{ccccc}

Way & Shot & $AP$ & $AP_{50}$ & $AP_{75}$ \\
\hline

\hline
\rowcolor{Gray1} 1 & 1 & 44.0 & 68.6 & 45.7 \\
\rowcolor{Gray2} 1 & 5 & 46.5 & 71.9 & 48.3 \\
\rowcolor{Gray1} 2 & 1 & 39.6 & 61.0 & 41.3 \\
\rowcolor{Gray2} 2 & 5 & 45.2 & 69.7 & 47.0 \\
\rowcolor{Gray1} 5 & 1 & 31.9 & 49.1 & 33.1 \\
\rowcolor{Gray2} 5 & 5 & 42.8 & 65.7 & 44.7 \\
\end{tabular}
\end{center}
\caption{Experimental results on FSVOD-500 val set of our model under different few-shot evaluation settings.}
\label{table:setting}
\vspace{-0.2in}
\end{table}

\begin{figure*}[!t]
\centering
%\fbox{\rule{0pt}{2.5in} \rule{1.0\linewidth}{0pt}}
\includegraphics[width=0.98\linewidth]{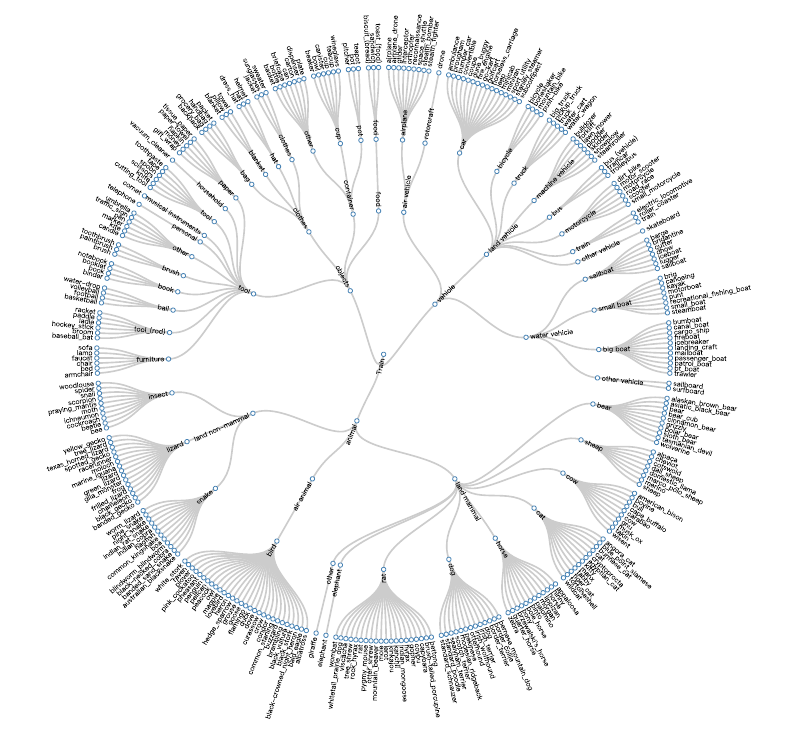}
\vspace{-0.2in}
\caption{Class hierarchy of FSVOD train set.}
\label{fig:train}
\vspace{-0.2in}
\end{figure*}

\begin{figure*}[!t]
\centering
%\fbox{\rule{0pt}{2.5in} \rule{1.0\linewidth}{0pt}}
\includegraphics[width=0.98\linewidth]{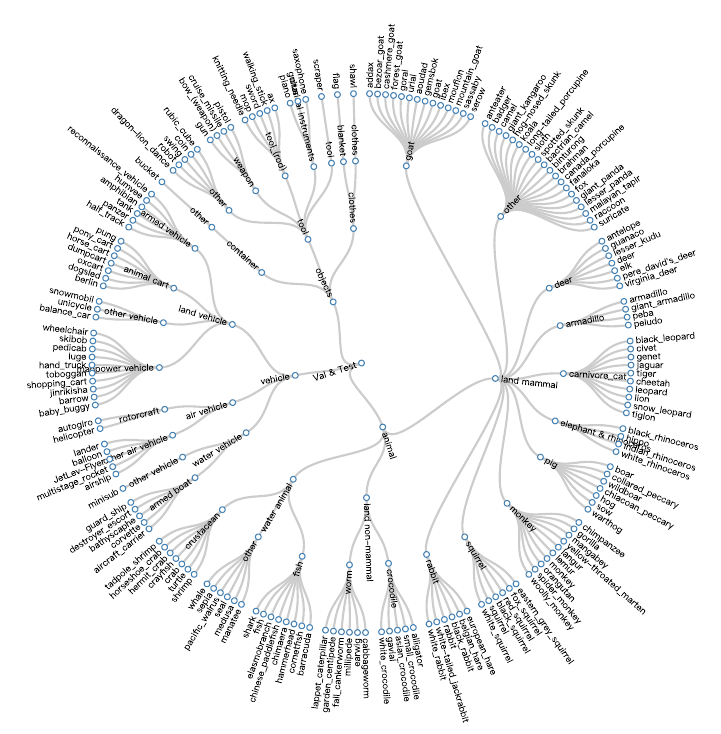}
\vspace{-0.2in}
\caption{Class hierarchy of FSVOD val and test sets.}
\label{fig:test}
\vspace{-0.2in}
\end{figure*}

\section{Full Dataset Hierarchy}

The full dataset hierarchy of FSVOD is shown in Figure~\ref{fig:train} (train set) and Figure~\ref{fig:test} (val and test sets).

\section{Object Localization in Massive Videos}

In the main paper, we propose a common realistic problem: \textit{Given a bunch of videos, how can we index and localize all novel objects of interest as video clips?.}

The practical solution is to detect objects in these videos and index/localize frames based on the detection results. Specifically, if there is a detection prediction for the target class, we index/localize this video frame.

The fully-supervised methods (\eg, object detection and multiple object tracking) can not solve this problem, because the interested objects can belong to arbitrary classes.

The single object tracking tasks can not solve this problem, because the video is massive and arbitrary, while the single object tracking requires the per-video annotated template for the first frame. Furthermore, the interested class may occur in discrete video clips and there are possibly multiple objects for the target class. Thus the single object tracking methods cannot handle these realistic cases.

The image-level few-shot learning tasks (\eg, few-shot image/video classification) can not solve this problem, because the interested objects is probably very small, while the image classification cannot properly represent small objects.

The few-shot image object detection cannot properly solve this problem, because its methods are specifically designed for still images without the consideration for the temporal information. %As shown in %Table~\ref{table:index}, the few-shot image object detection methods have low F1 score because of its massive false positive predictions and thus the precision is also very low.

The video object detection based methods are better than image detection based method because of their better detection results.
The multi-object tracking methods significantly improve the precision/recall/F1 performance thanks to the tube-based tracking.
Note these methods are all adapted for few-shot learning.
Our method has the best performance thanks to our tube proposal network and temporal matching strategy.

{\small
\bibliographystyle{splncs04}
\bibliography{main}
}

\end{document}